\documentclass[nohyperref]{article}

\usepackage{microtype}
\usepackage{graphicx}
\usepackage{subfigure}
\usepackage{booktabs} % for professional tables
\usepackage{amsfonts}
\usepackage{enumitem}
\usepackage{multirow}
\usepackage{footmisc}

\newlength\savewidth
\newcommand\shline{\noalign{\global\savewidth\arrayrulewidth
                            \global\arrayrulewidth 1.5pt}%
                   \hline
                   \noalign{\global\arrayrulewidth\savewidth}
                   }

\usepackage{hyperref}

\usepackage[accepted]{icml2023}

\usepackage{amsmath}
\usepackage{amssymb}
\usepackage{mathtools}
\usepackage{amsthm}

\usepackage[capitalize,noabbrev]{cleveref}

\theoremstyle{plain}

\theoremstyle{definition}

\theoremstyle{remark}

\usepackage[textsize=tiny]{todonotes}

\icmltitlerunning{Do We Really Need Graph Neural Networks for Traffic Forecasting?}

\begin{document}

\twocolumn[
\icmltitle{Do We Really Need Graph Neural Networks for Traffic Forecasting?}



\icmlsetsymbol{equal}{*}

\begin{icmlauthorlist}
\icmlauthor{Xu Liu}{nus}
\icmlauthor{Yuxuan Liang}{nus}
\icmlauthor{Chao Huang}{hku}
\icmlauthor{Hengchang Hu}{nus}
\icmlauthor{Yushi Cao}{ntu}
\icmlauthor{Bryan Hooi}{nus}
\icmlauthor{Roger Zimmermann}{nus}
\end{icmlauthorlist}

\icmlaffiliation{nus}{National University of Singapore}
\icmlaffiliation{hku}{University of Hong Kong}
\icmlaffiliation{ntu}{Nanyang Technological University}

\icmlcorrespondingauthor{Xu Liu}{liuxu@comp.nus.edu.sg}
\icmlcorrespondingauthor{Yuxuan Liang}{yuxliang@outlook.com}

\icmlkeywords{Machine Learning, ICML}

\vskip 0.3in
]


% This command actually creates the footnote in the first column listing the affiliations and the copyright notice.

\printAffiliationsAndNotice{}  % leave blank if no need to mention equal contribution

\begin{abstract}
    Spatio-temporal graph neural networks (STGNN) have become the most popular solution to traffic forecasting. While successful, they rely on the message passing scheme of GNNs to establish spatial dependencies between nodes, and thus inevitably inherit GNNs' notorious inefficiency. Given these facts, in this paper, we propose an embarrassingly simple yet remarkably effective spatio-temporal learning approach, entitled SimST. Specifically, SimST approximates the efficacies of GNNs by two spatial learning techniques, which respectively model local and global spatial correlations. Moreover, SimST can be used alongside various temporal models and involves a tailored training strategy. We conduct experiments on five traffic benchmarks to assess the capability of SimST in terms of efficiency and effectiveness. Empirical results show that SimST improves the prediction throughput by up to 39 times compared to more sophisticated STGNNs while attaining comparable performance, which indicates that \emph{GNNs are not the only option for spatial modeling in traffic forecasting}. 
\end{abstract}

\section{Introduction}
In recent years, urban traffic forecasting has emerged as one of the most important components of Intelligent Transportation Systems. Given historical traffic observations (e.g., traffic speed, flow) collected from sensors on road networks, the task focuses on predicting future traffic trends for each sensor, which provides insights for improving urban planning and traffic management \cite{zheng2014urban}. Spatio-Temporal Graph Neural Networks (STGNNs) have become the de facto most popular tool for traffic forecasting, in which sequential models such as Temporal Convolution Networks (TCNs) or Recurrent Neural Networks (RNNs) are applied for modeling temporal dependencies \cite{yu2018spatio,pan2019urban,wu2020connecting,lan2022dstagnn}, and Graph Neural Networks (GNNs) \cite{kipf2017semi,defferrard2016convolutional} are utilized to capture spatial correlations among different locations.
\vspace{-0.2em}

After revisiting the STGNNs proposed in recent years, we find that they mostly focus on enhancing the spatial learning modules (i.e., GNNs) with either complex aggregation rules or sophisticated layers to improve predictive performance. While successful, we note that they rely heavily on GNNs for performing the message passing step and thus inevitably inherit GNNs' notorious inefficiencies, especially when the graphs are large or with dense connections \cite{chen2021unified,zhang2022graph}. As a concrete example, the commonly used adaptive adjacency matrix is built by matrix multiplication of two node embedding tables, producing a fully-connected graph \cite{wu2019graph,bai2020adaptive,han2021dynamic}. During feature aggregation, such a fully-connected structure leads to \textit{quadratic} computational complexity w.r.t.~the number of sensors. Consequently, the scalability challenges of GNNs hinder the deployment of STGNNs in large-scale and real-time traffic forecasting systems that are latency-bound and require fast inference.
\vspace{-0.2em}

Although there are plenty of efforts in the graph domain to improve the efficiency of GNNs, such as via graph simplification \cite{hamilton2017inductive,chen2018fastgcn,chiang2019cluster,zeng2020graphsaint,zheng2020robust}, the related studies in STGNNs are still scarce to the best of our knowledge, as such simplification usually leads to information loss and performance degradation \cite{wu2020connecting}. Given these facts and inspired by recent progress in eliminating GNNs for node classification \cite{zhang2022graph,tian2022nosmog}, we may ask: \textbf{\emph{can we remove GNNs to trim down the explosive complexity while still remaining competitive in traffic forecasting accuracy? }}
\vspace{-0.5em}

\paragraph{Present Work}
To answer this question, we first demonstrate the functionalities of GNNs as follows. The superior performance of GNNs stems from its structure-aware exploitation of graphs: (1) for each node in the graph, a single GNN layer is used to first aggregate features from nodes' neighbors and then transform the aggregated representations via a feed-forward network, and (2) by stacking multiple layers, the hidden representations of nodes receive messages from long-distance neighbors.

In this work, we propose two spatial learning modules to \textit{approximate} the above efficacies of GNNs without requiring message passing, reducing the time complexity to \textit{linear}. (1) \textit{Local Proximity Modeling.} We take a local view and fragment the traffic network to build an \textit{ego-graph} for each node, which is constructed by incorporating historical observations of the node's neighbors. Then an MLP is applied to transform the observations to the hidden states at each time step. (2) \textit{Global Correlation Learning.} Inspired by recommender systems that learn user embeddings to reflect user behavioral similarities \cite{he2017neural,zhang2019deep}, we propose to use sensor embeddings to represent sensors' inherent properties and collaboratively capture spatial relationships between \textit{arbitrary sensor pairs} in a data-driven manner. Compared with stacking a number of GNN layers to enlarge the receptive field and build long-range spatial dependencies, our module sets up direct connections between nodes and thereby learns the global correlations.

In practice, GNNs can be used alongside various temporal models, such as GRU \cite{chung2014empirical}, WaveNet \cite{oord2016wavenet}, and Transformer \cite{vaswani2017attention}, for spatio-temporal learning. Analogously, we empirically find that our proposed spatial modules are agnostic to the temporal encoders and work in a plug-and-play manner. Moreover, we devise a new training strategy for SimST to further boost performance, which increases sample diversity during batch formation and enhances generalization. 

Integrating the above components, we propose a \underline{Sim}ple yet effective \underline{S}patio-\underline{T}emporal learning approach termed \textbf{SimST}, which \emph{trims the quadratic cost and performs on par with STGNNs in empirical studies}. Our contributions are summarized as follows.
\vspace{-1em}
\begin{itemize}[leftmargin=*]
    \item We for the first time demonstrate that GNNs are not the only choice for spatial modeling in traffic forecasting, by presenting a simple yet admirable method called SimST.
    
    \item Despite its simplicity, SimST achieves remarkable empirical results in terms of throughput and accuracy against sophisticated state-of-the-art STGNNs: SimST is significantly more efficient than baselines, with up to \textbf{39$\times$} higher inference throughput, while performing on par with them.
    
    \item We conduct ablation and case studies to promote a better understanding of our method and motivate future research to rethink the importance of GNNs in traffic forecasting.
    % smooth optimization landscape, faster convergence speed, and better generalization
    % We find SimST also has the advantages of low GPU memory usage, strong scalability, and fast convergence in the experiments.
\end{itemize}

\section{Related Work}
Traffic forecasting is a crucial application in smart city efforts. In recent years, STGNNs have become the most widely used tools for predicting traffic \cite{cao2020spectral,liu2022contrastive,chen2021z,chen2022tamp}. Generally, they integrate GNNs with either RNNs or TCNs to capture the spatial and temporal dependencies in traffic data. For example, DCRNN \cite{li2018diffusion} considered traffic flow as a diffusion process and combined a novel diffusion convolution with GRU. Other efforts \cite{pan2019urban,fang2021spatial,liu2022msdr} were also based on RNNs. To improve training speed and enjoy parallel computation, plenty of works \cite{wu2019graph,wu2020connecting,li2021spatial} replaced RNNs with dilated causal convolution.

\begin{figure*}[t]
  \centering
  \includegraphics[width=0.9\textwidth]{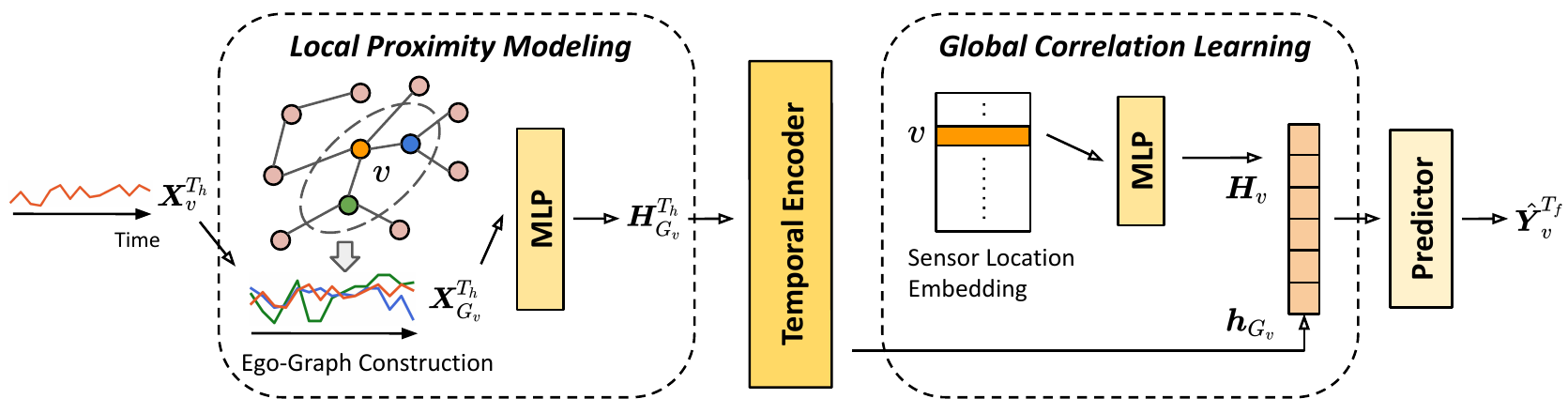}
  \vspace{-1em}
  \caption{An overview of SimST from the perspective of a node $v$ with $T_h$-step history. Our approach is essentially a single-node architecture (each node is handled individually by the model), which differs from existing STGNNs that require the entire graph as input.}
  \label{fig:framework}
  \vspace{-1em}
\end{figure*}

A fundamental problem for using GNNs in traffic modeling is how to establish a graph structure. The mainstream approach to define such structures is using either a predefined and sparse matrix constructed from the road network distances between sensors \cite{yu2018spatio,li2018diffusion,song2020spatial}, or an adaptive and dense matrix that records the pairwise relationships between nodes \cite{wu2019graph,bai2020adaptive,han2021dynamic,choi2022graph}. However, researchers have noted that the predefined matrix is heuristic and does not reflect the genuine dependencies between nodes, which degrades model performance \cite{wu2019graph,bai2020adaptive}. Models applying the adaptive matrix generally achieve superior performance. Though successful, the adaptive matrix discards the sparsity of the graph and incurs a quadratic computational cost, making it hard to deploy for latency-constrained or large-scale traffic applications. Other methods that use attention mechanisms for spatial learning \cite{zheng2020gman} also suffer from quadratic time complexity. In this study, we propose SimST to alleviate the inefficiency issue by eliminating the inefficient GNN component in the model.
% The most striking difference between STGNNs and the proposed SimST is that we do not apply message passing to spatial learning, which brings efficiency benefits.

% \citet{song2020spatial, han2021dynamic} incorporate the time dimension into graph structure, so as to build spatio-temporal graphs for jointly capturing spatio-temporal characteristics.

\section{Preliminary}
\paragraph{Traffic Forecasting}
Let $G = (\mathcal{V}, \mathcal{E})$ represent a directed sensor graph with $|\mathcal{V}|$ nodes and $|\mathcal{E}|$ edges, and $\boldsymbol{X}^T \in \mathbb{R}^{|\mathcal{V}| \times T \times F}$ denote the features of all nodes from time step 1 to $T$, where $F$ is the feature dimension. The features usually consist of a target attribute (e.g., traffic speed) and other auxiliary information, such as time of day \cite{wu2019graph}. Following common settings \cite{li2018diffusion,wu2019graph}, a weighted adjacency matrix $\boldsymbol{A} \in \mathbb{R}^{|\mathcal{V}| \times |\mathcal{V}|}$ is applied to describe the graph topology, where $\boldsymbol{A}_{ij} =  \exp(-\frac{{\rm dist}(v_i, v_j)^2}{s^2})$ if ${\rm dist}(v_i, v_j) \leqslant r$ else $\boldsymbol{A}_{ij} = 0$, ${\rm dist}(v_i, v_j)$ denotes the road network distance between sensors $v_i$ and $v_j$, $s$ is the standard deviation of distances, and $r$ is a threshold for sparsity \cite{shuman2013emerging}. The non-zero entries in $\boldsymbol{A}$ form the set of edges in $\mathcal{E}$.

In traffic forecasting, we aim to learn a neural network $\mathbf{\Theta}$ to predict the target attribute in future $T_f$ steps based on $T_h$ historical observations over the sensor graph:
\begin{align}
    G, \boldsymbol{X}^{T_h}
    \xrightarrow{\boldsymbol{\Theta}} \hat{\boldsymbol{Y}}^{T_f}
\end{align}
where $\boldsymbol{X}^{T_h} \in \mathbb{R}^{|\mathcal{V}| \times T_h \times F}$ indicates the observations and $\hat{\boldsymbol{Y}}^{T_f} \in \mathbb{R}^{|\mathcal{V}| \times T_f \times 1}$ is the predictions. A prediction loss, e.g., mean absolute error, is utilized to train the neural network:
\begin{align}
    \mathcal{L}(\mathbf{\Theta}) = \frac{1}{|\mathcal{V}|} \sum_{v_i \in \mathcal{V}} |\hat{\boldsymbol{Y}}_{v_i}^{T_f} - \boldsymbol{Y}_{v_i}^{T_f}|
\end{align}

\paragraph{Graph Neural Networks}
Although various forms of GNNs exist, our work refers to the conventional message passing scheme \cite{kipf2017semi} that is widely adopted in traffic forecasting. Specifically, message passing functions in two ways: aggregating one-hop neighbors to model local correlations, and stacking layers to build long-range relationships. Formally, the node representations $\boldsymbol{H}^l$ at the $l$-th layer of GNNs are learned by first aggregating messages from node neighbors $\mathcal{N}(v)$, and then transforming the representations via feed-forward propagation:
\begin{align}
    \boldsymbol{H}^l = \sigma(\tilde{\boldsymbol{A}}\boldsymbol{H}^{l-1}\boldsymbol{W}^l)
\end{align}
where $\tilde{\boldsymbol{A}}$ denotes the normalized adjacency matrix, $\boldsymbol{W}^l$ denotes the learnable weights at the $l$-th layer, and $\sigma$ is the activation function (e.g., ReLU).

\section{Methodology}
As a typical spatio-temporal data mining task, traffic forecasting requires modeling both spatial and temporal aspects. In this section, we start by describing how spatial correlations are set up by the proposed spatial learning modules in Section \ref{sec:4.1}. We then introduce three temporal encoders to capture the temporal dependencies and a predictor for generating the future in Section \ref{sec:4.2}. Finally, we present a node-based batch sampling method for training SimST in Section \ref{sec:4.3}. Figure \ref{fig:framework} depicts the architecture of SimST.

\subsection{Spatial Learning Modules}
\label{sec:4.1}
The inefficiency of GNNs has been extensively discussed within the graph domain \cite{chiang2019cluster,chen2021unified,zhang2022graph}. Relying on GNNs to set up spatial correlations via message passing, STGNNs also suffer from flagrant inefficiencies. In this work, we propose two simple yet effective alternatives to approximate the functionalities of GNNs and thus allow each node to exploit spatial information without message passing.

\paragraph{Local Proximity Modeling}
A single GNN layer is capable of aggregating neighborhood features in a non-Euclidean structure. To consider such a locality characteristic without using a GNN layer, we propose to first perform graph fragmentation to generate an \emph{ego-graph} for each node. The features of each ego-graph are the $T_h$ steps historical time series of the center node and its one-hop neighbors. Multi-hop neighbors can also be incorporated if needed, but we found that incorporating more neighbors may suffer from overfitting in experiments (see Section \ref{sec:5.4.3}). We then transform the raw features at each time step to their hidden representations via a multi-layer perceptron (MLP).

Specifically, for each node $v$, we consider its top-$k$ one-hop neighbors based on the weights in the normalized matrix $\tilde{\boldsymbol{A}}$, where $\tilde{\boldsymbol{A}} = \tilde{\boldsymbol{D}}^{-\frac{1}{2}} (\boldsymbol{A} + \boldsymbol{I}) \tilde{\boldsymbol{D}}^{-\frac{1}{2}}$ and $\tilde{\boldsymbol{D}}$ is the degree matrix of $\boldsymbol{A} + \boldsymbol{I}$. We also consider the neighbors in the reversed direction, i.e., the entries in $\boldsymbol{A}^{T}$, so that the model can perceive the impact from forward and backward neighbors. Let $\mathcal{N}^{1}_f(v)$ and $\mathcal{N}^{1}_b(v)$ denote the selected one-hop neighbors in two directions. We construct the feature matrix $\boldsymbol{X}^{T_h}_{G_v} \in \mathbb{R}^{T_h \times (2k+3)}$ of an ego-graph $G_{v}$ during period $T_h$:
\begin{align}
    \boldsymbol{X}^{T_h}_{G_v} = \text{COMBINE} (\boldsymbol{X}^{T_h}_{v}, 
    \{\boldsymbol{X}^{T_h}_{v_{f}} : v_f \in \mathcal{N}^1_f(v) \}, \nonumber
    \\
    \{\boldsymbol{X}^{T_h}_{v_{b}} : v_b \in \mathcal{N}^1_b(v) \}, \boldsymbol{X}^{T_h}_{avg_f}, \boldsymbol{X}^{T_h}_{avg_b})
\end{align}
where the last two terms are the average histories of all one-hop neighbors in two directions, allowing the center node to learn more about its local context. Note that the execution of building ego-graphs can be completed in data preprocessing, so it has no influence on model training and inference. Then, we apply an MLP to transform the raw features in $\boldsymbol{X}^{T_h}_{G_v}$ and generate the representations $\boldsymbol{H}^{T_h}_{G_v} \in \mathbb{R}^{T_h \times D_m}$, where $D_m$ is the model hidden dimension. Compared with a GNN layer, our module replaces message aggregation (i.e., weighted sum) operation with ego-graph construction that can be finished in the data preprocessing stage.

\paragraph{Global Correlation Learning}
Representing users with unique embeddings and learning with these embeddings to capture similarities in behavior among users, is a fundamental technique in recommender systems \cite{he2017neural,zhang2019deep}. Along a similar line, we argue that the traffic situation of a sensor is largely influenced by its specific position on a road, e.g., on the mainline or on a ramp. This is an intrinsic, time-invariant property of the sensor: hence, we propose to represent it using static sensor location embeddings. During end-to-end training, the embedding table collaboratively learns spatial relationships between \textit{arbitrary node pairs} in a data-driven manner. Apart from capturing global correlations, this module can serve as a complement to the local proximity modeling function, which is particularly useful when the connections in $\boldsymbol{A}$ are noisy \cite{wu2019graph,bai2020adaptive}. A detailed case study is conducted in Section \ref{sec:5.5}, where we show that the learned sensor embeddings contain significant information for sensor relevance reasoning.

Compared with stacking multiple GNN layers to enlarge the receptive field and establish long-range correlations, our proposed module uses embedding similarity to reflect the relationships between arbitrary sensor pairs, and thus further considers global knowledge. Moreover, the connections between sensors are built in a direct way, which keeps our module from over-squashing \cite{alon2021bottleneck}. As for implementation, we randomly initialize a low-dimensional embedding for each sensor, leading to an embedding table $\boldsymbol{E} \in \mathbb{R}^{|\mathcal{V}| \times D_n}$, where $D_n$ is the node embedding size. To fuse the embeddings with model hidden representations, we apply an MLP to map $\boldsymbol{E}$ to $\boldsymbol{H} \in \mathbb{R}^{|\mathcal{V}| \times D_m}$.

\paragraph{Complexity Comparison}
We provide a complexity comparison between the spatial learning modules here. As GCN \cite{kipf2017semi} is the widely adopted form of GNNs in spatio-temporal models, we take it as a baseline. Suppose a $L$-layer GCN with fixed hidden features $D_m$ is applied. For STGNNs applying the predefined adjacency matrix, the time complexity is $O(L|\mathcal{E}|D_m + L|\mathcal{V}|D^2_m)$. When applying the adaptive adjacency matrix that boosts performance, the complexity becomes $O(L|\mathcal{V}|^2D_m + L|\mathcal{V}|D^2_m)$. In SimST, the complexity of proximity modeling is $O(|\mathcal{V}|R)$, where $R$ is the average degree of nodes. The complexity of correlation learning is $O(|\mathcal{V}|D_nD_m)$. It is obvious that SimST has lower complexity and is capable of supporting forecasting applications on large-scale sensor networks.

\subsection{Temporal Encoder \& Predictor}
\label{sec:4.2}
There are several viable options for building the temporal dependencies of a node's history. To demonstrate that SimST can be generalized to various temporal models, we incorporate three basic and popular backbones in this study: GRU \cite{chung2014empirical} (RNN-based), WaveNet \cite{oord2016wavenet} (TCN-based), and Transformer \cite{vaswani2017attention}. The representations $\boldsymbol{H}^{T_h}_{G_v} \in \mathbb{R}^{T_h \times D_m}$ are the input of the temporal encoder. For GRU and WaveNet, they compress the temporal dimension to 1, generating the output $\boldsymbol{h}_{G_v} \in \mathbb{R}^{D_m}$, while for Transformer, the resulting representation still has the same shape as $\boldsymbol{H}^{T_h}_{G_v}$. We only take the last time step as the final output. 

Moreover, since the standard Transformer is aware of all input time steps during self-attention and the traffic status at the current step is in fact not conditioned on its future, we follow WaveNet to introduce causality into the attention mechanism, leading to the design of the Causal Transformer. This ensures that the model cannot violate the temporal order of inputs and such causality can be easily implemented by masking the specific entries in the attention map. Finally, we concatenate the spatio-temporal summary $\boldsymbol{h}_{G_v} \in \mathbb{R}^{D_m}$ of node $v$ with its location embedding $\boldsymbol{H}_v \in \mathbb{R}^{D_m}$, which is then fed into the predictor (implemented as an MLP) for generating the forecasting results of node $v$: $\hat{\boldsymbol{Y}}_{v}^{T_f} = \text{MLP}(\boldsymbol{h}_{G_v} || \boldsymbol{H}_v)$.

\begin{table*}[t]
    \centering
    \small
    \tabcolsep=0.85mm
    \caption{Test MAE, RMSE, and MAPE results averaged over all predicted time steps on five datasets. We bold the best results and underline the second best results. $^*$ denotes the improvement of SimST over the best baseline is statistically significant at level 0.05.}
    \label{tab:performance}
    \vspace{0.5em}
    \begin{tabular}{l|ccc|ccc|ccc|ccc|ccc}
        \shline
        \multirow{2}{*}{Method} & \multicolumn{3}{c|}{PeMSD4} & \multicolumn{3}{c|}{PeMSD7} & \multicolumn{3}{c|}{PeMSD8} & \multicolumn{3}{c|}{LA} & \multicolumn{3}{c}{BAY} \\ \cline{2-16} 
        & MAE & RMSE & MAPE & MAE & RMSE & MAPE & MAE & RMSE & MAPE & MAE & RMSE & MAPE & MAE & RMSE & MAPE \\
        \hline \hline
        HA & 38.03 & 59.24 & 27.88\% & 45.12 & 65.64 & 24.51\% & 34.86 & 59.24 & 27.88\% & 4.16 & 7.80 & 13.00\% & 2.88 & 5.59 & 6.80\% \\
        VAR & 24.54 & 38.61 & 17.24\% & 50.22 & 75.63 & 32.22\% & 19.19 & 29.81 & 13.10\% & 5.28 & 9.06 & 12.50\% & 2.24 & 3.96 & 4.83\% \\
        \hline
        DCRNN & 22.39 & 34.93 & 15.19\% & 23.41 & 36.66 & 9.98\% & 16.62 & 25.95 & 10.60\% & 3.14 & 6.28 & \underline{8.65\%} & 1.63 & 3.65 & 3.70\% \\
        STGCN & 21.59 & 33.83 & 15.49\% & 26.12 & 41.43 & 11.92\% & 17.41 & 26.72 & 11.78\% & 3.39 & 6.79 & 9.34\% & 1.88 & 4.30 & 4.28\% \\
        ASTGCN & 21.58 & 33.76 & 14.71\% & 25.77 & 39.41 & 11.67\% & 17.91 & 27.34 & 11.36\% & 3.57 & 7.19 & 10.32\% & 1.86 & 4.07 & 4.27\% \\
        GWNET & 19.33 & \textbf{30.73} & 13.18\% & 20.65 & \underline{33.47} & 8.84\% & \textbf{14.98} & \textbf{23.75} & 9.75\% & \textbf{3.06} & \textbf{6.10} & \textbf{8.38\%} & \underline{1.58} & 3.54 & \underline{3.59\%} \\
        STSGCN & 21.19 & 33.65 & 13.90\% & 24.26 & 39.03 & 10.21\% & 17.13 & 26.80 & 10.96\% & 3.32 & 6.66 & 9.06\% & 1.79 & 3.91 & 4.06\% \\
        AGCRN & 19.83 & 32.26 & \underline{12.97\%} & 20.69 & 34.19 & 8.86\% & 15.95 & 25.22 & 10.09\% & 3.19 & 6.41 & 8.84\% & 1.62 & 3.61 & 3.66\% \\
        STGODE & 20.84 & 32.82 & 13.77\% & 22.99 & 37.54 & 10.14\% & 16.81 & 25.97 & 10.62\% & 4.73 & 7.60 & 11.71\% & 1.77 & \textbf{3.33} & 4.02\% \\
        STGNCDE & \underline{19.21} & 31.09 & \textbf{12.76\%} & \underline{20.53} & 33.84 & 8.80\% & 15.45 & 24.81 & 9.92\% & 3.58 & 6.84 & 9.91\% & 1.68 & 3.66 & 3.80\% \\
        GMSDR & 20.49 & 32.13 & 14.15\% & 22.27 & 34.94 & 9.86\% & 16.36 & 25.58 & 10.28\% & 3.21 & 6.41 & 8.76\% & 1.69 & 3.80 & 3.74\% \\
        \hline
        SimST-WN & 19.56 & 31.24 & 13.38\% & 20.81 & 33.94 & 8.87\% & 15.55 & 24.70 & 10.17\% & 3.19 & 6.35 & 8.97\% & 1.60 & 3.55 & 3.66\% \\
        SimST-CT & 19.32 & 30.98 & 13.35\% & 20.63 & 33.71 & \underline{8.70\%} & 15.13 & \underline{24.18} & \underline{9.73\%} & \underline{3.11} & \underline{6.19} & 8.73\% & 1.59 & 3.50 & 3.61\% \\
        SimST-GRU & \textbf{19.19} & \underline{30.92} & 13.13\% & \textbf{20.14$^*$} & \textbf{33.34} & \textbf{8.46\%} & \underline{14.99} & 24.26 & \textbf{9.66\%} & 3.16 & 6.29 & 8.82\% & \textbf{1.57} & \underline{3.47} & \textbf{3.58\%} \\
        \shline
    \end{tabular}
    \vspace{-1em}
\end{table*}
% HA, VAR
% DCRNN, STGCN, ASTGCN, GWNET, STSGCN, AGCRN, STGODE, STGNCDE, GMSDR
% Candidate: GMAN, MTGNN, STFGNN, Z-GCNET, DMSTGCN, TAMP-S2GCN, DSTAGNN

\subsection{Node-based Batch Sampling for Training SimST}
\label{sec:4.3}
When training an STGNN, the input data are usually a four-dimensional tensor $\boldsymbol{X}^{T_h} \in \mathbb{R}^{B \times |\mathcal{V}| \times T_h \times F}$, where $B$ denotes the set batch size. That said, nodes are strictly bound together and the batch sampling unit is the entire graph. However, directly applying this graph-based sampling to SimST leads to two issues. First, since SimST is a single-node architecture, the actual batch size $B^*$ for SimST is $B \times |\mathcal{V}|$. For example, there are 883 nodes in the PeMSD7 dataset \cite{song2020spatial} and usually $B$ is set to 64, then $B^* =\ $ 56,512. Such a large number reduces the noise in the gradient estimation and leads to a degenerated generalization \cite{keskar2017large}. Also, when $|\mathcal{V}|$ is large, it can easily cause GPU out-of-memory problems. Second, we argue that the graph-based sampling method significantly diminishes sample diversity when forming mini-batches. For example, the traffic flow of all nodes from 8 to 9 a.m.~on January 26 will always appear in the same batch.

To tackle these issues, we propose to remove the restriction on nodes and change the batch sampling unit from a graph to a single node. Note that the same node at different time periods is viewed as different instances. Consequently, the input to SimST becomes $\boldsymbol{X}^{T_h} \in \mathbb{R}^{B^* \times T_h \times F}$, where $B^* \ll B \times |\mathcal{V}|$. The benefits are two-fold. First, our solution provides flexibility to choose $B^*$, which can be disproportionate to $|\mathcal{V}|$ and thus allows for smaller values like 256. This property enhances generalization \cite{keskar2017large}, and allows SimST to have low GPU memory costs and to easily scale to large graphs. Second, our approach provides more sample diversity when forming mini-batches, which yields a generalization improvement \cite{ash2020deep}.

\section{Experiments}
\subsection{Experimental Setup}
\paragraph{Datasets}
We conduct experiments on commonly used traffic benchmarks. The details are presented in Table \ref{tab:dataset}. All traffic readings are aggregated into 5-minute windows, resulting in 288 data points per day. Note that the values in traffic flow datasets are generally much larger than those in speed-related datasets. Following \cite{li2018diffusion,song2020spatial,choi2022graph}, we use the 12-step historical data to predict the next 12 steps. Z-score normalization is applied to the input data for fast training. We build the adjacency matrix of each dataset by using road network distances with a thresholded Gaussian kernel \cite{shuman2013emerging}. The threshold $r$ is set to 0 for PeMSD4, PeMSD7, PeMSD8, and 0.1 for LA and BAY. More information is provided in Appendix \ref{app:a}.

\paragraph{Baselines}
We compare SimST with the following baselines. Historical Average (HA) \cite{pan2012utilizing} and Vector Autoregression (VAR) \cite{toda1991vector} are traditional methods. DCRNN \cite{li2018diffusion}, STGCN \cite{yu2018spatio} , ASTGCN \cite{guo2019attention}, GWNET \cite{wu2019graph}, STSGCN \cite{song2020spatial}, and AGCRN \cite{bai2020adaptive} are well-known STGNNs. We also incorporate methods that are published recently: STGODE \cite{fang2021spatial}, STGNCDE \cite{choi2022graph}, GMSDR \cite{liu2022msdr}.

\paragraph{Implementation Details} 
SimST is implemented with PyTorch 1.12. There are three variants of SimST based on the applied temporal encoders. We describe the specific configurations for different variants in Appendix \ref{app:b}. The following settings are the same for all variants. We train our method via the Adam optimizer with an initial learning rate of 0.001 and a weight decay of 0.0001. We set the maximum epochs to 150 and use an early stop strategy with a patience of 20. The batch size is 1,024. The embedding size $D_n$ is 20. The top-$k$ nearest neighbors are set to 3 for LA and BAY, and 0 for PeMSD4, PeMSD7, and PeMSD8 (due to limited available edges). For baselines, we run their codes based on the recommended configurations if their accuracy is not known for a dataset. If known, we use their officially reported accuracy. Experiments are repeated five times with different seeds on an NVIDIA RTX A6000 GPU.

\begin{table}[t]
    \tabcolsep=0.8mm
    \caption{Dataset statistics.}
    \label{tab:dataset}
    \vspace{0.5em}
    \centering
    \small
    \begin{tabular}{l|cccc}
        \shline
         Dataset & \#Node & \#Edge & \#Instance &  Attribute \\
        \hline \hline
         PeMSD4 \cite{song2020spatial} & 307 & 338 & 16,992 & Flow \\
         PeMSD7 \cite{song2020spatial} & 883 & 865 & 28,224 & Flow \\
         PeMSD8 \cite{song2020spatial} & 170 & 276 & 17,856 & Flow \\
         LA \cite{li2018diffusion} & 207 & 1,515 & 34,272 & Speed \\
         BAY \cite{li2018diffusion} & 325 & 2,369 & 52,116 & Speed \\
        \shline
    \end{tabular}
    \vspace{-1em}
\end{table}

\paragraph{Evaluation Metrics} We adopt three common metrics in forecasting tasks to evaluate the model performance, including mean absolute error (MAE), root mean squared error (RMSE), and mean absolute percentage error (MAPE). For efficiency measurement, the commonly used floating point operations per second (FLOPS) cannot reflect the “real” running speed of methods because it ignores the effects of parallelization. Hence, we apply the metric of throughput per second (TPS) in this study, which indicates the average number of samples the network can process in one second. We provide a wall-clock time comparison in Appendix \ref{app:c}.

\begin{table*}[t]
    \centering
    \small
    \tabcolsep=0.9mm
    \caption{A efficiency comparison between five best-performing baselines and SimST variants. Param denotes the number of learnable parameters and the magnitude of parameters is Kilo ($10^3$). TPS is measured on the validation set during inference. We bold the highest TPS. The $\Delta$ column means the TPS improvements of SimST-WN over all the other methods.}
    \label{tab:efficiency}
    \vspace{0.5em}
    \begin{tabular}{l|ccc|ccc|ccc|ccc|ccc}
        \shline
        \multirow{2}{*}{Method} & \multicolumn{3}{c|}{PeMSD4} & \multicolumn{3}{c|}{PeMSD7} & \multicolumn{3}{c|}{PeMSD8} & \multicolumn{3}{c|}{LA} & \multicolumn{3}{c}{BAY} \\ \cline{2-16} 
        & Param & TPS & $\Delta$ & Param & TPS & $\Delta$ & Param & TPS & $\Delta$ & Param & TPS & $\Delta$ & Param & TPS & $\Delta$ \\
        \hline \hline
        GWNET & 303 & 1,876 & $4.15\times$ & 314 & 522 & $5.65\times$ & 300 & 3,258 & $3.75\times$ & 301 & 2,825 & $3.30\times$ & 303 & 1,738 & $3.52\times$ \\
        AGCRN & 749 & 1,559 & $4.99\times$ & 755 & 445 & $6.62\times$ & 747 & 1,733 & $7.06\times$ & 748 & 1,604 & $5.82\times$ & 749 & 1,466 & $4.17\times$ \\
        STGODE & 715 & 552 & $14.09\times$ & 733 & 226 & $13.04\times$ & 710 & 721 & $16.97\times$ & 709 & 682 & $13.69\times$ & 713 & 532 & $11.49\times$ \\
        STGNCDE & 377 & 217 & $35.85\times$ & 388 & 76 & $38.79\times$ & 374 & 362 & $33.79\times$ & 375 & 251 & $37.19\times$ & 377 & 204 & $29.97\times$ \\
        GMSDR & 880 & 267 & $29.14\times$ & 2253 & 119 & $24.77\times$ & 423 & 465 & $26.31\times$ & 641 & 300 & $31.12\times$ & 923 & 238 & $25.68\times$ \\
        \hline
        SimST-CT & 147 & 4,597 & $1.69\times$ & 159 & 1,656 & $1.78\times$ & 144 & 7,887 & $1.55\times$ & 145 & 5,969 & $1.56\times$ & 148 & 4,041 & $1.51\times$ \\
        SimST-GRU & 130 & 6,167 & $1.26\times$ & 142 & 2,268 & $1.30\times$ & 127 & 10,494 & $1.17\times$ & 128 & 7,766 & $1.20\times$ & 131 & 5,440 & $1.12\times$ \\
        SimST-WN & 167 & \textbf{7,780} & - & 179 & \textbf{2,948} & - & 164 & \textbf{12,232} & - & 165 & \textbf{9,335} & - & 168 & \textbf{6,113} & - \\
        \shline
    \end{tabular}
    \vspace{-1em}
\end{table*}

\subsection{Performance Comparison}
In this section, we conduct a model comparison between SimST variants and state-of-the-art STGNNs on five real-world traffic benchmarks. The three variants of SimST are SimST-GRU, SimST-WaveNet (WN), and SimST-Causal Transformer (CT). According to the empirical results in Table \ref{tab:performance}, all variants of SimST achieve competitive performance on the three evaluation metrics of all datasets, which validates the effectiveness of SimST and indicates SimST can generalize to various architectures of temporal models. For comparison among SimST variants: note that for fairness we ensure that the variants have a similar number of parameters (around 150k). SimST-GRU generally achieves comparable performance to state-of-the-art baselines, i.e., GWNET and STGNCDE. SimST-CT also attains competitive accuracy, with little difference from SimST-GRU and GWNET. We notice that while Transformers are generally considered a more powerful model, our results show that a simple GRU works well for traffic prediction. In summary, our results provide strong support for our claim that GNNs are not the only option for spatial modeling in traffic forecasting. Moreover, considering the very different design of SimST compared to other state-of-the-art methods, we hope that this will inspire follow-on studies in graph-less designs.

Additionally, we find that: (1) STGNNs approaches surpass HA and VAR by a large margin due to their greater learning capacity. (2) The capability of methods that apply the adaptive matrix, such as GWNET and AGCRN, has been overlooked. They have achieved similar performance to the recently proposed approaches.

\subsection{Efficiency Comparison}
In this part, we select the top-performing approaches and compare their efficiency with SimST variants. It can be seen from Table \ref{tab:efficiency} that all SimST variants are significantly more efficient than baselines, thanks to their simple model architecture and linear time complexity w.r.t.~$|\mathcal{V}|$. Specifically, comparing TCN-based models, SimST-WN has around 45\% fewer parameters than GWNET but achieves 3.3$\times$ -- 5.6$\times$ higher TPS. For RNN-based methods, SimST-GRU is 3.7$\times$ -- 6.1$\times$ and 19$\times$ -- 26$\times$ faster than AGCRN and GMSDR, respectively. STGODE and STGNCDE are the recently published methods that are based on neural ordinary differential equations. Though achieving competitive performance, they suffer from significant inefficiency problems. For example, SimST-WN has comparable performance to them but is 11$\times$ -- 17$\times$ and 30$\times$ -- 39$\times$ faster. Among the variants, SimST-WN achieves the highest TPS due to WaveNet's superior parallelism capability. SimST-GRU also achieves good TPS results, which we attribute to an optimized implementation in the PyTorch library. See Appendix \ref{app:c} for more results on efficiency comparison.

\begin{figure}[!b]
  \centering
  \vspace{-2em}
  \includegraphics[width=\linewidth]{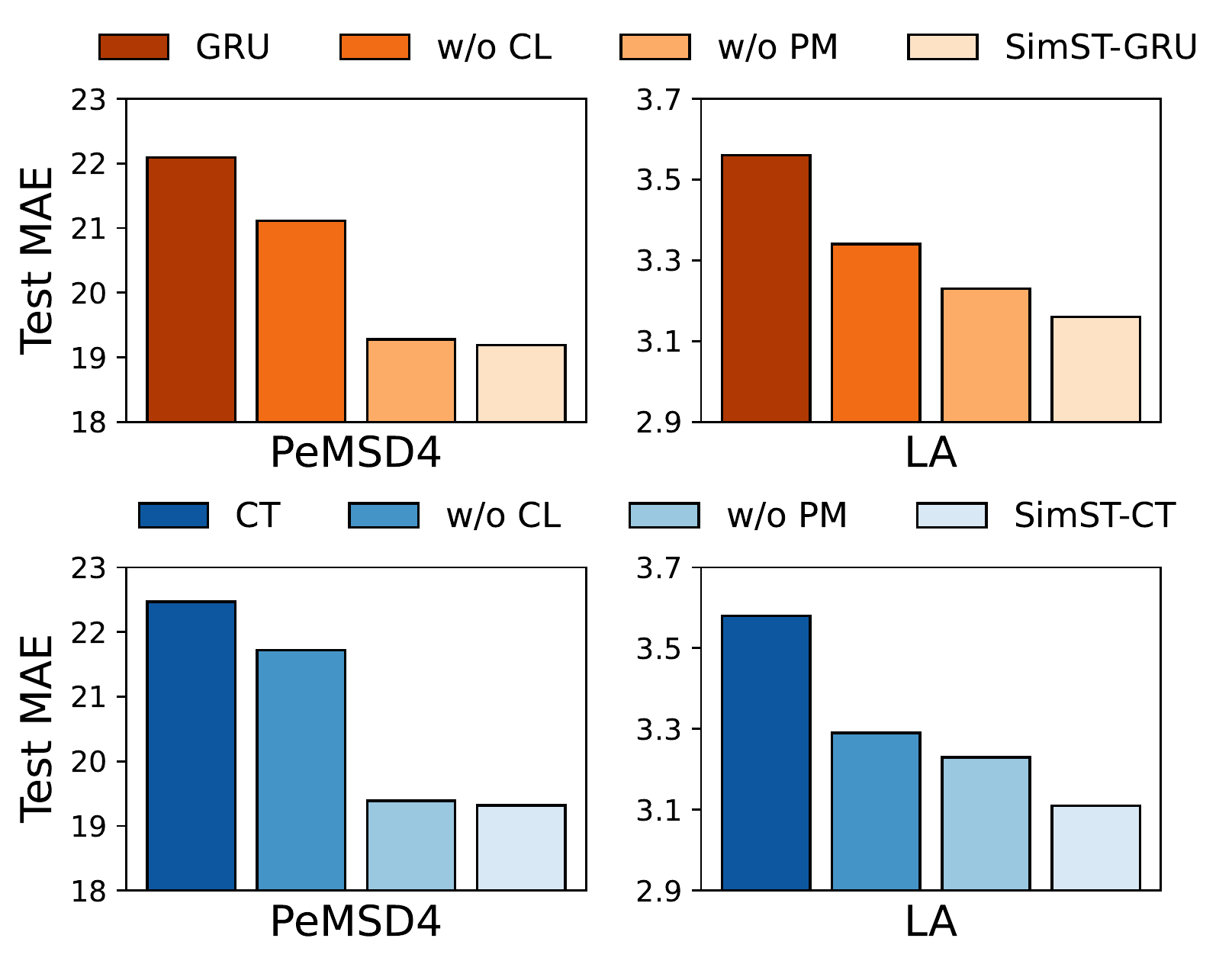}
  \vspace{-2em}
  \caption{Effects of two spatial learning modules: correlation learning (CL) and proximity modeling (PM).}
  \label{fig:spatial}
  \vspace{-1em}
\end{figure}

\begin{figure}[t]
  \centering
  \includegraphics[width=\linewidth]{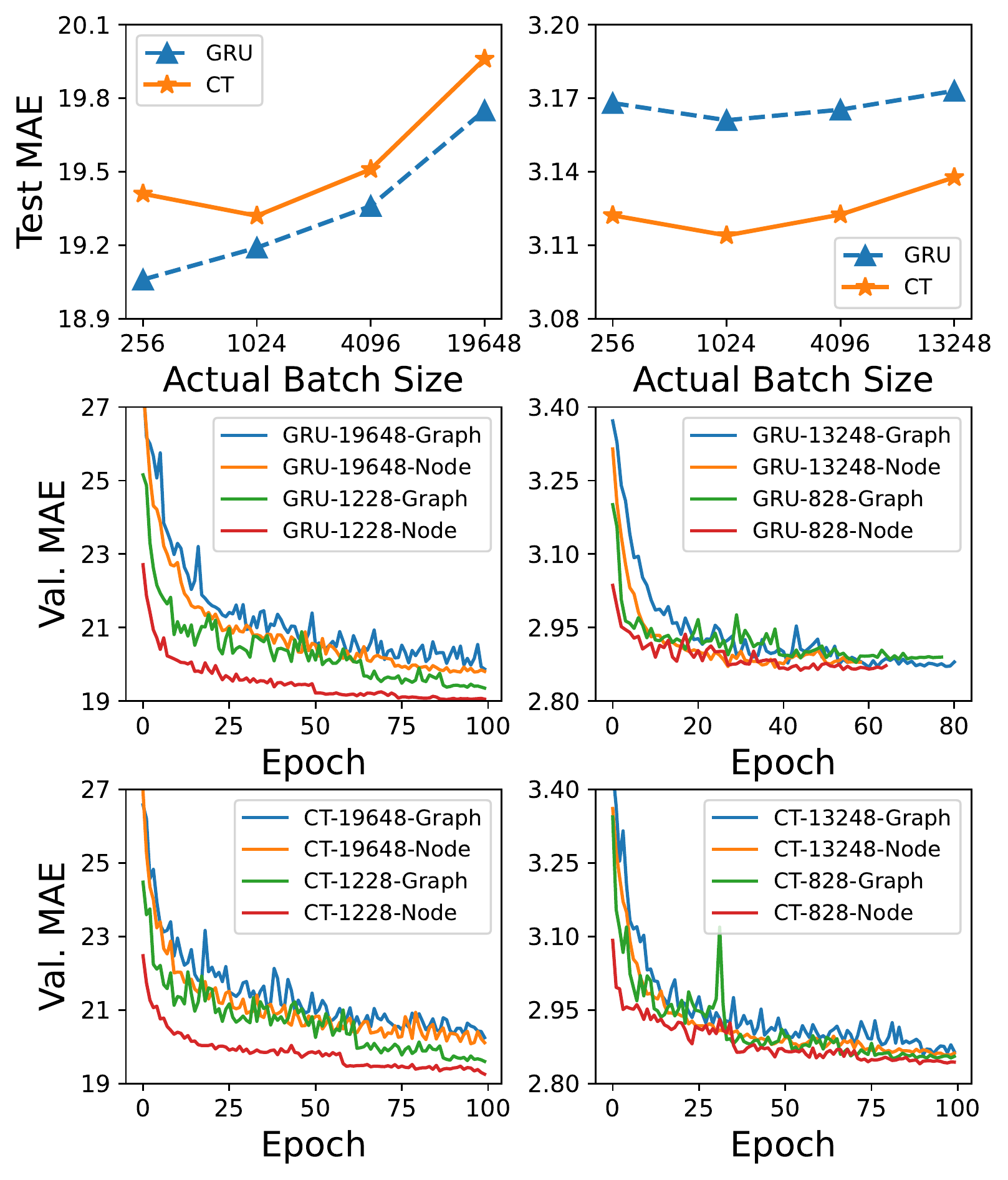}
  \vspace{-2em}
  \caption{Effects of the node-based batch sampling strategy on PeMSD4 (left column) and LA (right column) datasets. For simplicity, here we denote SimST-GRU as GRU and SimST-CT as CT. In the first row, the largest values in the x-axis are computed by $|\mathcal{V}| \times$ 64 (the common batch size used in STGNNs \protect\cite{wu2019graph,bai2020adaptive,choi2022graph}). In the second and third rows, the number in the legend denotes $B^*$.}
  \label{fig:strategy}
  \vspace{-1em}
\end{figure}

\begin{figure}[t]
  \centering
  % \vspace{0.7em}
  \includegraphics[width=\linewidth]{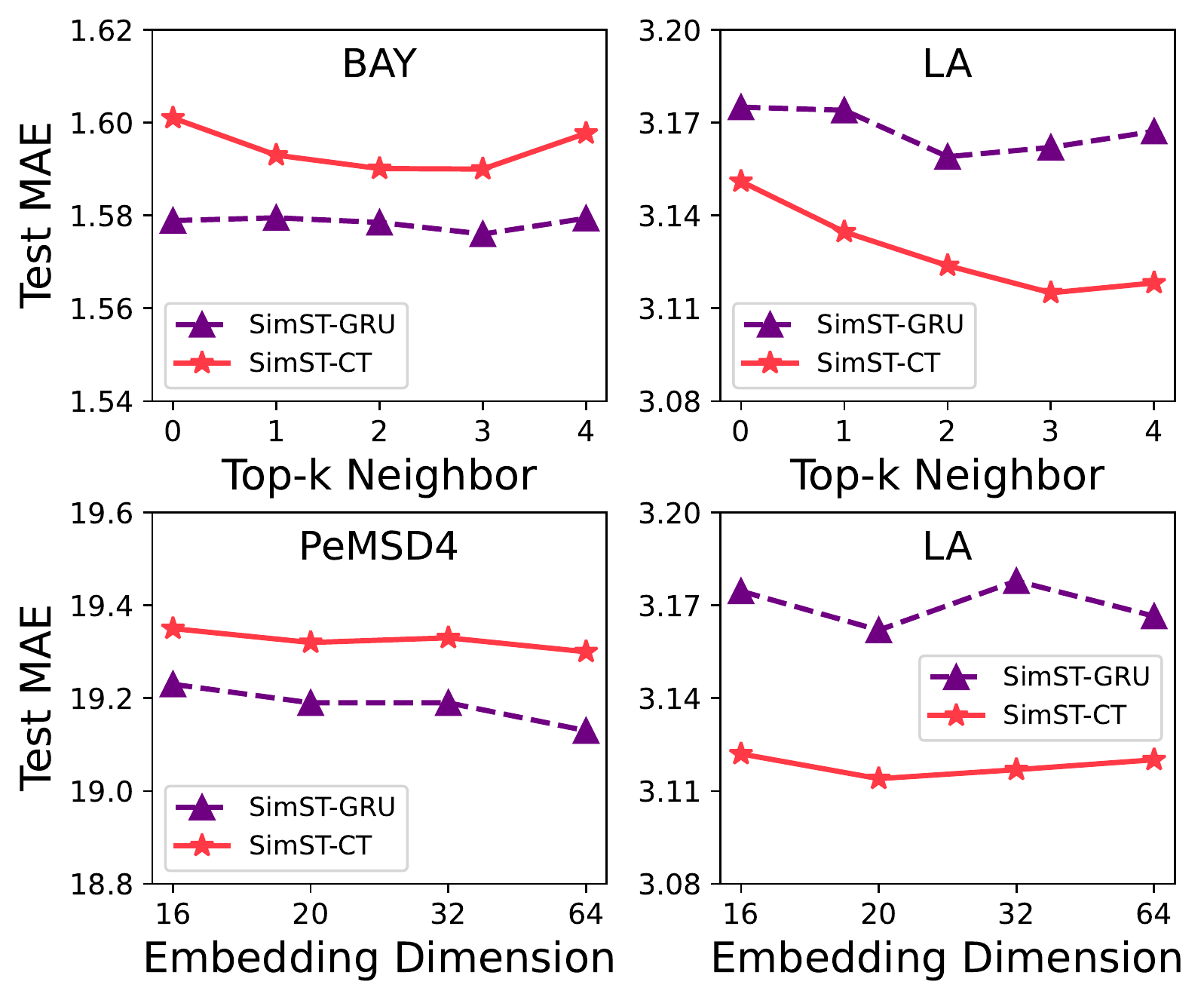}
  \vspace{-2em}
  \caption{Effects of the top-$k$ neighbors on BAY and LA datasets, and sensor embedding dimension on PeMSD4 and LA datasets.}
  \label{fig:parameter}
  \vspace{-1em}
\end{figure}

\subsection{Ablation Study}
We have demonstrated the effectiveness and efficiency of SimST. Next, we select SimST-GRU and SimST-CT as the representatives (due to their preferable performance) and conduct a series of ablation and case studies on one flow-based dataset PeMSD4, and one speed-related dataset LA.

\subsubsection{Effects of spatial learning modules}
To study the effects of two spatial modules, we consider the following three settings for comparisons: (1) \textbf{w/o Correlation Learning (CL)}: we turn off the spatial correlation learning module. (2) \textbf{w/o Proximity Modeling (PM)}: for each node, we only input the histories of itself. (3) \textbf{CT/GRU}: we do not perform spatial learning. The results are shown in Figure \ref{fig:spatial}. First, we find that removing CL leads to significant degradation of MAE for both SimST-GRU and SimST-CT on both datasets, revealing the great importance of building global correlations under the SimST framework. Second, the benefit of PM is marginal on PeMSD4. This is because the average degree of nodes in PeMSD4 is only 1.1, so the available neighbor information is very limited. In contrast, the average degree of nodes in the LA dataset is 7.3, where PM affects performance more. Third, we notice that removing the CL module has a greater impact on the PeMSD4 dataset than on the LA dataset. The reason is that when neighbor information is scarce, the CL module takes a greater responsibility to supplement neighbor knowledge. Conversely, when there are more neighbors, the influence of CL on performance is less significant.

\subsubsection{Effects of Node-based Batch Sampling}
We examine the effects of the node-based sampling method from two aspects. First, we show the influence of the batch size number in the first row of Figure \ref{fig:strategy}. Generally, both SimST-GRU and SimST-CT reach their best performance when setting $B^*$ to 1,024 on the two datasets, revealing the necessity of applying a small $B^*$. The negative influence of using a large batch size such as 19,648, is more significant on the PeMSD4 dataset. Besides, we find that SimST-CT only occupies around 2GB of memory on our GPU across all datasets. In contrast, the memory cost of STGNNs usually grows as $|\mathcal{V}|$ increases. For example, on the dataset with the lowest $|\mathcal{V}| = 170$ and the highest $|\mathcal{V}| = 883$, GWNET requires around 3GB and 11GB of memory, respectively.

Second, we present the effects of different batch-forming approaches, i.e., graph-based sampling and node-based sampling by drawing the convergence curves in the second and third rows of Figure \ref{fig:strategy}. Note that we ensure fair comparisons by setting the same $B^*$ in experiments. In Figure \ref{fig:strategy}, it can be seen that node-based batch sampling surpasses the counterpart. Concretely, node-based sampling is only slightly better than the graph-based method when a large $B^*$ is applied (e.g., 19,648/13,248). This is because the sample diversity of graph-based sampling is rich enough at this time. As $B^*$ decreases to small values (e.g., 1,228/828), which is necessary to improve performance, the impact of sample diversity becomes significant, i.e., generalization performance is greatly improved, especially on PeMSD4. Also, we find that the curves of node-based sampling are generally more stable than the graph-based method.

\begin{figure*}[t]
  \centering
  \includegraphics[width=\textwidth]{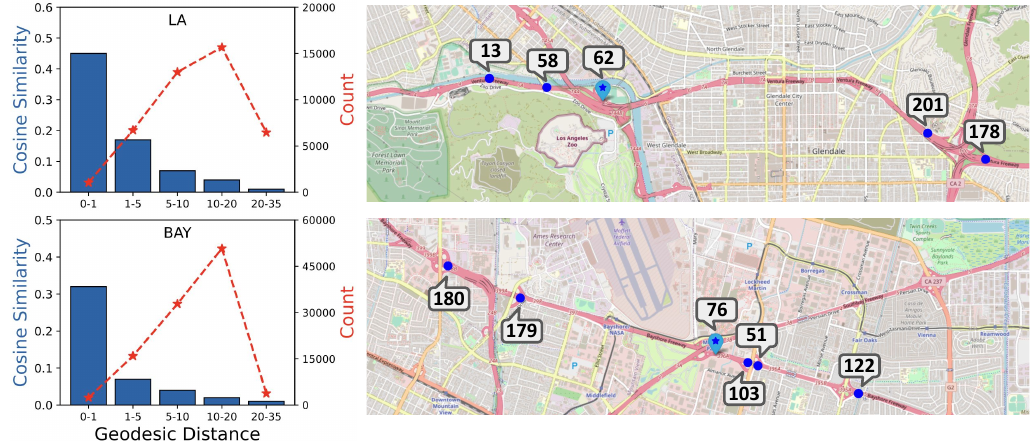}
  \vspace{-2em}
  \caption{A case study of learned sensor embeddings. The Count axis on the left part means the number of sensor pairs within the corresponding ranges, and the trend is indicated by red dotted line. The sum of the counts is $|\mathcal{V}|^2$. On the right part, the star icons represent the anchor sensor, the circle icons denote the selected similar sensors, and the roads colored in red are the highways in California, USA.}
  \label{fig:case}
  \vspace{-0.5em}
\end{figure*}

\subsubsection{Hyperparameter Study}
\label{sec:5.4.3}
In Figure \ref{fig:parameter}, first row, we show the effects of the number of top-$k$ one-hop neighbors. We replace PeMSD4 with the BAY dataset because the average degree of nodes is only 1.1 in PeMSD4. From the figure, we observe that the performance gradually improves when $k$ increases from 0 to 3. But the performance worsens when we set $k$ to 4, indicating a potential overfitting issue. Next, we assess the effects of the sensor embedding dimension size in the second row of Figure \ref{fig:parameter}. We find that both SimST-GRU and SimST-CT are not sensitive to the changes in the dimension on both datasets. The results also suggest that a small dimension such as 16, is sufficient to learn correlations between nodes.

\subsection{Case Study}
\label{sec:5.5}
In this section, we explore further to understand \textit{what exactly the sensor embeddings have learned} through a case study that makes connections between embedding similarities and real-world sensor locations. We apply SimST-CT on LA and BAY since only these two datasets provide sensors' coordinates. Concretely, we first compute pairwise cosine similarities between sensor embeddings, i.e., for $v_i, v_j \in \mathcal{V}, \text{similarity} = \frac{E_{v_i} \cdot E_{v_j}}{\left\|E_{v_i}\right\|_{2}\left\|E_{v_j}\right\|_{2}}$, and pairwise geodesic distances between sensors based on their coordinates. Then we calculate the average cosine similarities within the ranges of 0--1, 1--5, 5--10, 10--20, and 20--35 kilometers of all sensors, leading to a statistical overview shown in Figure \ref{fig:case} (left part). It can be seen that cosine similarity becomes smaller when the geodesic distance gets larger. This result obeys Tobler’s First Law of Geography, i.e., near things are more related than distant things.

Next, we look at an example by selecting sensor 62 in LA as the anchor node, which is located at the intersection of two highways. We also visualize its top similar neighbors based on cosine similarity in Figure \ref{fig:case} (upper right). We find that (1) sensors 13 and 58 are the nearby neighbors, which are also included in the adjacency matrix. (2) sensors 201 and 178 are distant nodes. They are considered similar sensors because they are also located at intersections, and thus share similar traffic patterns to the anchor. The situation in the BAY dataset is similar to LA (shown in the lower right of Figure \ref{fig:case}), so we do not elaborate further.

\subsection{Discussion \& Future Work}
We find that SimST-GRU and SimST-CT perform worse than the state-of-the-art STGNNs (i.e., STGNCDE and GWNET) when we turn off the node-based batch sampling strategy. This result indicates that the proposed two spatial learning modules can only approximate the efficacies of GNNs, and the training strategy of SimST is an indispensable component to obtain comparable performance to the state-of-the-art. In addition, we note that when the neighbor information is relatively rich, such as with LA, SimST-CT is slightly below the state-of-the-art. Therefore, we plan to address this performance gap by utilizing pretrained models and knowledge distillation \cite{zhang2022graph}. Lastly, our findings in this work are currently limited to traffic forecasting, while we would like to expand SimST to other spatio-temporal applications such as air quality prediction.

\section{Conclusion}
In this paper, we propose SimST, a simple, effective, and efficient spatio-temporal learning method for traffic forecasting. It approximates GNNs with two proposed spatial learning modules, involves a node-based batch sampling strategy, and is temporal-model-agnostic. Our study suggests that message passing is not the only effective way of modeling spatial relations in traffic forecasting; hence, we hope to spur the development of new model designs with both high efficiency and effectiveness in the community.
\clearpage

% We also would like to point out several potential future directions. Though SimST can generalize to newly added sensors easily, it still needs to be fine-tuned over the new nodes and is not in a fully inductive manner. A future study may focus on developing a completely inductive spatio-temporal model.

% \section*{Acknowledgements}

\bibliography{icml2023}
\bibliographystyle{icml2023}

%%%%%%%%%%%%%%%%%%%%%%%%%%%%%%%%%%%%%%%%%%%%%%%%%%%%%%%%%%%%%%%%%%%%%%%%%%%%%%%
%%%%%%%%%%%%%%%%%%%%%%%%%%%%%%%%%%%%%%%%%%%%%%%%%%%%%%%%%%%%%%%%%%%%%%%%%%%%%%%
% APPENDIX
%%%%%%%%%%%%%%%%%%%%%%%%%%%%%%%%%%%%%%%%%%%%%%%%%%%%%%%%%%%%%%%%%%%%%%%%%%%%%%%
%%%%%%%%%%%%%%%%%%%%%%%%%%%%%%%%%%%%%%%%%%%%%%%%%%%%%%%%%%%%%%%%%%%%%%%%%%%%%%%
\clearpage
\appendix
% \onecolumn

\section{Datasets}

\label{app:a}
We conduct experiments on the traffic datasets of PeMSD4\footnote{https://github.com/Davidham3/STSGCN\label{note1}}, PeMSD7\footref{note1}, PeMSD8\footref{note1}, LA\footnote{\label{note2}https://github.com/liyaguang/DCRNN} and BAY\footref{note2}.
\begin{itemize}[leftmargin=*]
    \item PEMS-04 \cite{song2020spatial}: The dataset refers to the traffic flow in San Francisco Bay Area. There are 307 sensors and the period ranges from Jan. 1 - Feb. 28, 2018.
    \item PEMS-07 \cite{song2020spatial}: This dataset involves traffic flow readings collected from 883 sensors, and the time range is from May 1 - Aug. 6, 2017. 
    \item PEMS-08 \cite{song2020spatial}: The dataset contains traffic flow information collected from 170 sensors in the San Bernardino area from Jul. 1 - Aug. 31, 2016.
    \item LA \cite{li2018diffusion}: The dataset records the traffic speed collected from 207 loop detectors in the highway of Los Angeles County, ranging from Mar. 1 - Jun. 27, 2012. 
    \item BAY \cite{li2018diffusion}: This dataset contains traffic speed information from 325 sensors in the Bay Area. It has 6 months of data ranging from Jan. 1 - Jun. 30, 2017.
\end{itemize}

The datasets are split into three parts for training, validation, and testing with a ratio of 6:2:2 on PeMSD4, PeMSD7, and PeMSD8, and 7:1:2 on LA and BAY. The statistics of data partition are in Table \ref{tab:partition}.

\begin{table}[h]
    \centering
    \tabcolsep=3mm
    \caption{Details of data partition.}
    \label{tab:partition}
    \vspace{0.5em}
    \begin{tabular}{l|ccc}
        \shline
         Dataset & \#Training & \#Validation & \#Testing \\
        \hline
         PeMSD4 & 10,172 & 3,375 & 3,376 \\
         PeMSD7 & 16,911 & 5,622 & 5,622 \\
         PeMSD8 & 10,690 & 3,548 & 3,549 \\
         LA & 23,974 & 3,425 & 6,850 \\
         BAY & 36,465 & 5,209 & 10,419  \\
        \shline
    \end{tabular} 
\end{table}

\section{Configuration of SimST Variants}
\label{app:b}
\begin{table}[H]
    \centering
    \small
    \tabcolsep=0.9mm
    \caption{Detailed configurations of different SimST variants. Dim: dimension. FFN: Feed-Forward Network.}
    \label{tab:config}
    \vspace{0.5em}
    \begin{tabular}{l|ccc}
        \shline
         Setting & SimST-WN & SimST-GRU & SimST-CT \\
        \hline
         Model Hidden Dim $D_m$ & 64 & 64 & 64 \\
         Kernel Size in WN & 3 & - & - \\
         Skip Connect Dim in WN & 64 & - & - \\
         Head Number in CT & - & - & 2 \\
         FFN Dim in CT & - & - & 128 \\
         Predictor Dim & 512 & 512 & 512 \\
         Layer Number & 3 & 2 & 2 \\
         Dropout Ratio & 0.1 & 0.1 & 0.1 \\
         Clip Norm & 5 & 5 & 5 \\
        \shline
    \end{tabular}
\end{table}

\section{Additional Performance Comparison}
\begin{figure*}[t]
  \centering
  \includegraphics[width=0.95\textwidth]{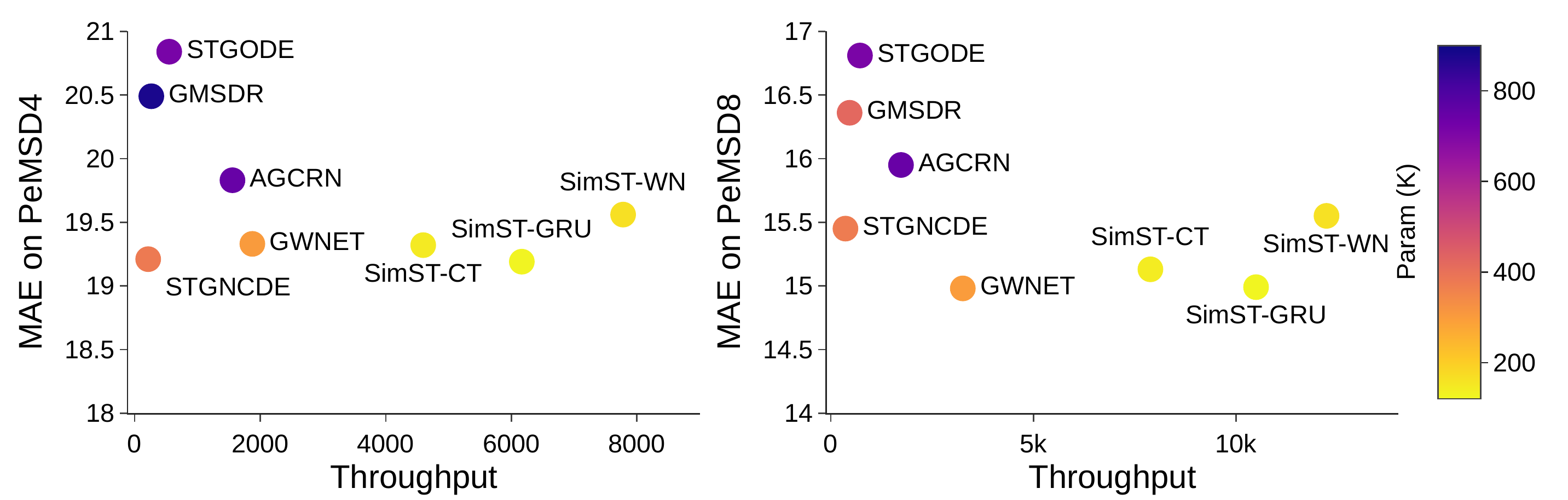}
  \vspace{-1em}
  \caption{A visual comparison between top-performing STGNNs and SimST variants on the popular PeMSD4 and PeMSD8 datasets. Throughput means the number of samples the network can process in a second during inference.}
  \label{fig:overall}
  \vspace{-1em}
\end{figure*}

In Figure \ref{fig:overall}, we provide a visual comparison of throughput, accuracy, and parameter size between top-performing STGNNs and SimST. It can be seen that SimST variants are substantially more efficient and use much fewer parameters than state-of-the-art STGNNs, while achieving comparable performance. Additionally, we notice that adding an auxiliary feature -- day of week can significantly improve the predictive accuracy, as shown in Table \ref{tab:addres}. The reason is that there is a significant difference in traffic patterns between weekdays and weekends. Therefore, this feature is a powerful prior knowledge for the model, and future research may consider incorporating this feature to enhance performance.

\section{Wall-Clock Time Comparison}
\label{app:c}
In Table \ref{tab:efficiency2}, we provide the training and inference time per epoch of top-performing baselines and SimST variants, to enable straightforward efficiency comparison. We observe that SimST variants have significantly shorter inference time, thanks to their linear complexity w.r.t the number of nodes and small parameter size. As for training efficiency, although SimST-GRU applies a small actual batch size $B^* = 1,024$, which is necessary to achieve better predictive performance, its training time is still comparable to the fastest two baselines GWNET and AGCRN. Furthermore, we notice that methods based on ordinary differential equations (STGODE and STGNCDE) require significant training and inference time, which largely affects their practical applications in the real world.

Moreover, since we need to perform ego-graph construction for the proximity modeling module, we provide details of the time spent here based on our current implementation: 41.9 seconds on PeMSD4, 206.5 seconds on PeMSD7, 28.8 seconds on PeMSD8, 284.1 seconds on LA, and 716.1 seconds on BAY. Note that this construction process only need to run once, because such process is deterministic and we can store the processed data and load them during training and inference.

\section{Learning Curve Comparison}
\begin{figure*}[t]
  \centering
  \includegraphics[width=0.95\textwidth]{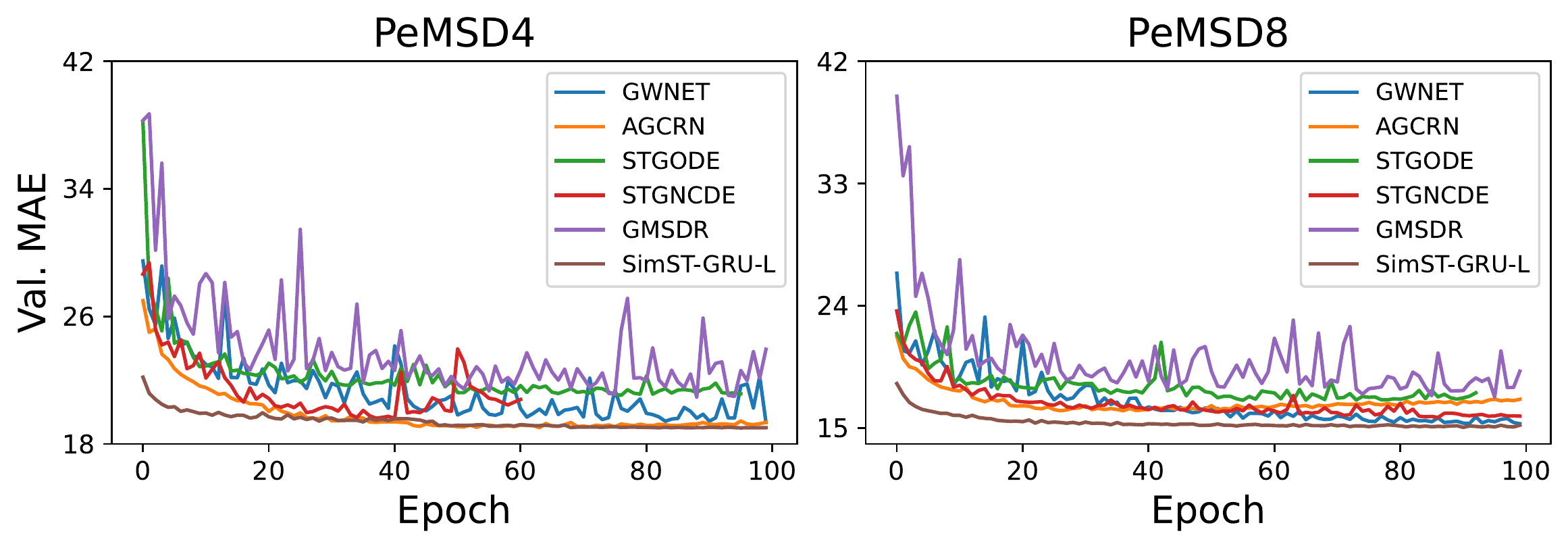}
  \vspace{-1em}
  \caption{Learning curve comparison between top-performing baselines and SimST-GRU-L on the validation set of PeMSD4 and PeMSD8.}
  \label{fig:curve}
  \vspace{-1em}
\end{figure*}

In this section, we first scale up the parameter size of SimST-GRU from 130k to 340k (denoted as SimST-GRU-L), to ensure our method has comparable parameters to the baselines. Then in Figure \ref{fig:curve}, we compare the learning curves between top-performing baselines and SimST-GRU-L. It can be seen that SimST-GRU-L exhibits very fast convergence rate: it drops to a small validation MAE within a few epochs, and almost converges in around 20 epochs. In addition, we observe that the standard deviation of SimST is generally smaller than those in baselines, indicating the stability provided by our approach.

\begin{table*}[t]
    \centering
    \small
    \tabcolsep=0.7mm
    \caption{Additional test results of SimST variants after adding the day of week feature to the inputs. We bold the best results.}
    \label{tab:addres}
    \vspace{0.5em}
    \begin{tabular}{l|ccc|ccc|ccc|ccc|ccc}
        \shline
        \multirow{2}{*}{Method} & \multicolumn{3}{c|}{PeMSD4} & \multicolumn{3}{c|}{PeMSD7} & \multicolumn{3}{c|}{PeMSD8} & \multicolumn{3}{c|}{LA} & \multicolumn{3}{c}{BAY} \\ \cline{2-16} 
        & MAE & RMSE & MAPE & MAE & RMSE & MAPE & MAE & RMSE & MAPE & MAE & RMSE & MAPE & MAE & RMSE & MAPE \\
        \hline \hline
        SimST-WN & 19.04 & 30.74 & 12.93\% & 19.88 & 32.94 & 8.39\% & 15.11 & 24.28 & 9.96\% & 3.10 & 6.21 & 8.88\% & 1.57 & 3.47 & 3.58\% \\
        SimST-GRU & 18.70 & 30.33 & \textbf{12.52\%} & \textbf{19.42} & \textbf{32.61} & \textbf{8.09\%} & 14.71 & 23.96 & 9.69\% & 3.08 & 6.16 & 8.84\% & \textbf{1.55} & \textbf{3.40} & \textbf{3.50\%} \\
        SimST-CT & \textbf{18.68} & \textbf{30.30} & 12.72\% & 19.70 & 32.73 & 8.25\% & \textbf{14.64} & \textbf{23.70} & \textbf{9.55\%} & \textbf{3.04} & \textbf{6.08} & \textbf{8.69\%} & 1.56 & 3.45 & 3.56\% \\
        % SimST* & 18.61 &  &  & 19.55 &  &  & 14.27 &  &  & 3.05 &  &  & 1.56 &  &  \\
        \shline
    \end{tabular}
    % \vspace{-1em}
\end{table*}

\begin{table*}[h]
    \centering
    \small
    \tabcolsep=2.1mm
    \caption{Efficiency comparison between five best-performing baselines and SimST variants. Train: training time (s) per epoch. Infer: inference time (s) per epoch. We bold the best results and underline the second best results.}
    \label{tab:efficiency2}
    \vspace{0.5em}
    \begin{tabular}{l|cc|cc|cc|cc|cc}
        \shline
        \multirow{2}{*}{Method} & \multicolumn{2}{c|}{PeMSD4} & \multicolumn{2}{c|}{PeMSD7} & \multicolumn{2}{c|}{PeMSD8} & \multicolumn{2}{c|}{LA} & \multicolumn{2}{c}{BAY} \\ \cline{2-11} 
        & Train & Infer & Train & Infer & Train & Infer & Train & Infer & Train & Infer  \\
        \hline \hline
        GWNET & \textbf{19.45} & 1.81 & \underline{101.32} & 10.79 & \textbf{11.46} & 1.10 & \textbf{30.63} & 1.22 & \underline{72.11} & 3.02 \\
        AGCRN & \underline{22.46} & 2.18 & \textbf{98.59} & 12.66 & 16.19 & 2.07 & 40.18 & 2.15 & \textbf{68.42} & 3.58 \\
        STGODE & 69.46 & 6.11 & 272.90 & 24.88 & 67.56 & 4.92 & 151.20 & 4.99 & 264.27 & 9.75 \\
        STGNCDE & 148.54 & 15.55 & 716.63 & 73.97 & 94.22 & 9.80 & 337.84 & 13.65 & 571.39 & 25.53 \\
        GMSDR & 80.63 & 12.70 & 311.61 & 47.32 & 50.96 & 7.71 & 170.74 & 11.52 & 328.18 & 22.05 \\
        \hline
        SimST-CT & 39.82 & 0.74 & 184.74 & 3.40 & 22.32 & 0.45 & 62.95 & 0.58 & 152.57 & 1.30  \\
        SimST-GRU & 23.76 & \underline{0.55} & 112.39 & \underline{2.48} & \underline{13.64} & \underline{0.34} & \underline{39.75} & \underline{0.45} & 94.08 & \underline{0.96} \\
        SimST-WN & 42.90 & \textbf{0.43} & 205.25 & \textbf{1.91} & 24.47 & \textbf{0.29} & 72.12 & \textbf{0.37} & 177.49 & \textbf{0.85} \\
        \shline
    \end{tabular}
    % \vspace{-1em}
\end{table*}

% \section{Visualization}
% To have a better sense of the capacity of our method, in this part, we randomly select two nodes and visualize the forecasting results of the SimST-CT model and the corresponding ground truths. 

% \begin{figure*}[h]
%   \centering
%   \includegraphics[width=\textwidth]{figures/vis_pred1.pdf}
%   % \vspace{-2em}
%   \caption{Visualization on the test set of PeMSD4.}
%   \label{fig:pred1}
% %   \vspace{-1em}
% \end{figure*}

% \begin{figure*}[h]
%   \centering
%   \includegraphics[width=\textwidth]{figures/vis_pred2.pdf}
%   % \vspace{-2em}
%   \caption{Visualization on the test set of PeMSD7.}
%   \label{fig:pred2}
% %   \vspace{-1em}
% \end{figure*}

% \begin{figure*}[h]
%   \centering
%   \includegraphics[width=\textwidth]{figures/vis_pred3.pdf}
%   % \vspace{-2em}
%   \caption{Visualization on the test set of PeMSD8.}
%   \label{fig:pred3}
% %   \vspace{-1em}
% \end{figure*}

%%%%%%%%%%%%%%%%%%%%%%%%%%%%%%%%%%%%%%%%%%%%%%%%%%%%%%%%%%%%%%%%%%%%%%%%%%%%%%%
%%%%%%%%%%%%%%%%%%%%%%%%%%%%%%%%%%%%%%%%%%%%%%%%%%%%%%%%%%%%%%%%%%%%%%%%%%%%%%%

\end{document}